\documentclass[letterpaper]{article}
\usepackage{aaai}
\usepackage{times}
\usepackage{helvet} 
\usepackage{courier} 
\usepackage{graphicx}
\usepackage{latexsym}
\usepackage{amsmath}
\usepackage{amssymb}
\usepackage{makeidx}
\usepackage{multicol}
\usepackage{tikz}
\usepackage{graphicx}
\usepackage[vlined,algoruled,linesnumbered]{algorithm2e}
\usepackage{subfigure}
\usepackage{dblfloatfix}

\newtheorem{definition}{Definition}
\newtheorem{proposition}{Proposition}

\newtheorem{lemma}{Lemma}
\newtheorem{proof}{Proof}

\newcommand{\PS}{\text{PS} }
\newcommand{\AN}{\texttt{AND} }
\newcommand{\ON}{\texttt{OR} }

\def\nnf{{\tt NNF}}
\def\cnf{{\tt CNF}}
\def\dnf{{\tt DNF}}
\def\mods{{\tt MODS}}
\def\Dnnf{{\tt DNNF}}
\def\sDnnf{{\tt SDNNF}}
\def\tDnnf{{\tt DNNF$_T$}}
\def\dnnf{{\tt d-NNF}}
\def\snnf{{\tt s-NNF}}
\def\fnnf{{\tt f-NNF}}
\def\dDnnf{{\tt d-DNNF}}
\def\sdDnnf{{\tt sd-DNNF}}
\def\bdd{{\tt BDD}}
\def\fbdd{{\tt FBDD}}
\def\obdd{{\tt OBDD}}
\def\oobdd{{\tt OBDD$_{<}$}}
\def\pi{{\tt PI}}
\def\ip{{\tt IP}}
\def\co{{\sf \bf CO}}
\def\cd{{\sf \bf CD}}

\def\opt{{\sf \bf OPT}}

\def\y{$\surd$}
\def\n{$\times$}
\def\fpt{$\oslash$}

\newcommand{\myproof}[1]{}

\newcommand{\proofTextInText}[1]{}

\begin{document}

\title{On the Complexity of Optimization Problems\\ based on Compiled NNF Representations}

\author{Daniel Le Berre \and Emmanuel Lonca \and Pierre
  Marquis\\
CRIL-CNRS, Universit\'e d'Artois, France\\
\tt{lastname@cril.fr}}

\maketitle


\bibliographystyle{aaai}

\begin{abstract}
  Optimization is a key task in a number of applications.  When the
  set of feasible solutions under consideration is of combinatorial
  nature and described in an implicit way as a set of constraints,
  optimization is typically {\sf NP}-hard. Fortunately, in many
  problems, the set of feasible
  solutions does not often change and is independent from the
  user's request. In such cases, compiling the set of constraints
  describing the set of feasible solutions during an off-line phase
  makes sense, if this compilation step renders computationally easier
  the generation of a non-dominated, yet feasible solution matching
  the user's requirements and preferences (which are only known at the
  on-line step). In this article, we focus on propositional
  constraints.  The subsets $L$ of the \nnf\ language analyzed in
  Darwiche and Marquis' knowledge compilation map are considered. A
  number of families $\mathcal{F}$ of representations of objective
  functions over propositional variables, including linear
  pseudo-Boolean functions and more sophisticated ones, 
  are considered. For each language
  $L$ and each family $\mathcal{F}$, the complexity of
  generating an optimal solution when the constraints are compiled
  into $L$ and optimality is to be considered w.r.t. a function from
  $\mathcal{F}$ is identified. 
\end{abstract}

\section{Introduction}

%

Many applications from AI and other domains amount to an optimization task 
(using e.g. the pseudo-Boolean optimization \cite{PBO} or MaxSat \cite{MAXSAT} representation).
However, when the set of feasible solutions under consideration is of combinatorial nature,
and described in an implicit way as a set of constraints, optimization
is typically intractable in the worst case.
Fortunately, in many problems, 
the set of feasible solutions does not often change and is independent
from the optimization criterion.  As a matter of example, consider the 
software dependency management problem, and
more precisely, the GNU/Linux package dependency management
problem \cite{mancinelli2006managing}. Constraints are of the form ``package $A$ in version $1$ requires
package $B$ in any version'' and ``package $A$ in version $2$ requires
packages $B$ and $C$ in any version'', ``both versions of package $A$ cannot be installed together''
and can be encoded by propositional formulae like $\phi = (A \Leftrightarrow (A_1 \vee A_2)) \wedge$
$(B \Leftrightarrow (B_1 \vee B_2)) \wedge$ $ (C \Leftrightarrow (C_1 \vee C_2)) \wedge$ $(A_1 \Rightarrow B)
\wedge$ $(A_2 \Rightarrow (B \wedge C)) \wedge$ $(\neg A_1 \vee \neg A_2)$.
Given the hard constraint $\phi$, an initial state ``package $B$ is installed in
version $1$'',  and more generally some user requirements ``install
package $A$'', a dependency solver must find (if possible) a set of packages to be
installed such that the constraints and the user requirements are fulfilled.
Such a decision problem is {\sf NP}-complete \cite{debian-asp-99,mancinelli2006managing}.
Considering in addition some user preferences about the  packages to be installed leads to an {\sf NP}-hard optimization 
problem for which several specific solvers have been designed recently \cite{opiumPaper,p2cudf,aspcud,packup}.

Clearly enough, in the software dependency management problem (as in many configuration
problems), all the available pieces of information do not play the same role: 
all users share the same constraints, as the dependencies between
packages do not depend on the user. What makes each user ``specific"
is the initial state of her system, and more generally her own requirements, as well as the preferences she can have over the
feasible solutions. Pursuing the toy example above, the dependency problem admits the two following
solutions: $\{A_1,A,B_{1},B\}, \{A_2,A,B_{1},B,C_{1},C\}$. A user who favors the least
changes between the initial and the final states would prefer the
first solution to the second one, whereas a user who favors the most recent packages
would make the other choice. 
In such cases, compiling the set of constraints describing the set of feasible 
solutions during an off-line phase makes sense, 
if this compilation step renders computationally easier
the generation of a non-dominated, yet feasible solution matching the user's requirements and preferences
(which are only known at the on-line step). 

In the following, we focus on propositional constraints.
We consider the languages $L$ analyzed in \cite{compilationMap} as target languages for knowledge
compilation. Each of those languages $L$ satisfies the conditioning transformation, i.e., for each of them, there exists a
polynomial-time algorithm which associates with each formula $\phi \in L$ and each consistent
term $\gamma$ representing a partial assignment a formula from $L$ representing the
conditioning $\phi \mid \gamma$ of $\phi$ by $\gamma$. This conditioning $\phi \mid \gamma$ is equivalent to
the most general consequence of $\phi \wedge \gamma$ which is independent of the variables occurring
in $\gamma$. Equivalently, it is the formula obtained by substituting in $\phi$ each occurrence of a variable $x$ of
$\gamma$ by the Boolean constant $\top$ (resp. $\bot$) if the polarity of $x$ in $\gamma$ is positive (resp. negative).
The fact that the conditioning transformation is tractable for each of those languages enables to take into account efficiently an initial state and 
 the user's requirements (on our running example, $\gamma = A \wedge B_1$) during the on-line phase.
 
We also consider a number of families $\mathcal{F}$ of representations of objective functions $f$ over propositional variables,
including linear pseudo-Boolean functions, and more sophisticated ones (like polynomial pseudo-Boolean 
functions \cite{boros2002pseudo}). Here, solutions correspond to propositional interpretations $\omega$ and criteria are
represented by propositional formulae $\phi_i$, and are thus of Boolean nature. We note $\phi_i(\omega) = 1$ when
$\omega$ is a model of $\phi_i$, otherwise $\phi_i(\omega) = 0$. 
The importance of a criterion $\phi_i$ is measured by the weight $w_i$ associated with it (a real number). 
Each weight $w_i$ expresses the penalty of satisfying $\phi_i$ (it is a cost when positive and a reward otherwise).
Each function $f$ is represented by a weighted base $\{(\phi_i, w_i) \mid i \in \{1, \ldots, n\}\}$, 
i.e., a finite multi-set of propositional formulae, where each formula $\phi_i$ is associated its weight and is also characterized
by the aggregator $\oplus$ used to combine the weights $w_i$. $\omega$ is feasible
when it satisfies the hard constraint $\phi \in L$; in such a case, $f(\omega)$ is defined as $f(\omega) = \oplus_{i=1}^n w_i . \phi_i(\omega)$.
Two aggregators are considered: a utilitarist one ($\oplus = \Sigma$) and an egalitarist one ($\oplus = \mathit{leximax}$).
An optimal solution $\omega^*$ is a feasible one which minimizes the value of $f$.
 
The contribution of the paper is a complexity landscape for the optimization problem in such a setting, i.e.,
the problem of determining an optimal solution (when it exists) given $\phi \in L$ and a weighted base from $\mathcal{F}$.
More precisely, for each language $L$, each family $\mathcal{F}$ of representations, and each of the two aggregators, 
one determines whether this problem can be solved in polynomial time or it cannot be solved in polynomial time unless
{\sf P = NP}. The problem is considered in the general case, and under the restriction when the cardinality of the weighted 
base is bounded.

%
%

The rest of the paper is organized as follows.
After some formal preliminaries, we focus on the family of linear representations of pseudo-Boolean objective functions, characterized by
weighted bases for which each $\phi_i$ is a literal. 
We show that \Dnnf\ and its subsets ~\cite{darwiche1999compiling,darwiche2001decomposable} 
are precisely the subsets $L$ of \nnf\ (among those identified in the knowledge compilation map)
for which the optimization problem can be solved in polynomial time. 
Especially, this problem is {\sf NP}-hard for all the other subsets in the general case.
Afterwards, we switch to the more general family of polynomial representations of pseudo-Boolean objective 
functions, i.e., when each $\phi_i$ is a term, and finally to the more general case when each $\phi_i$ is any \nnf\ formula.
For theses families, the optimization problem is {\sf NP}-hard for each language $L$, even under some strong restrictions
on the representations of the objective function $f$, except when $L =$ \dnf. Finally, we show that some additional
tractable cases can be reached (especially for the polynomial representations of $f$) provided that a preset number
of criteria is considered (i.e., the cardinality of the weighted base is bounded).



\section{Formal Preliminaries}


In the following, we consider a finite set of propositional
variables denoted $\PS$; we sometimes omit it in the notations (when it is not
ambiguous to do so).
$\bot$ is the Boolean constant
always false, while $\top$ is the Boolean constant
always true.
$l = \neg x$ with $x \in \PS$ is a negative literal, and
$l = x$ with $x \in \PS$ is a positive literal. 
 Boolean constants are also considered as positive literals.
If $l = \neg x$ (resp. $l = x$) 
then its complementary literal $\sim l$ is $\sim l = x$ (resp. $\sim l = \neg x$).
A complete assignment $\omega$ of variables in $\text{PS}$ is called an
interpretation: $\omega$ is a set of pairs $(v \in
\PS,\{0,1\})$ where each $v \in \PS$ is the first projection of
exactly one pair in $\omega$.  $\omega$ is also viewed as the canonical term
$\bigwedge_{(v, 0) \in \omega} \neg v \wedge \bigwedge_{(v, 1) \in \omega} v$.
$\Omega_{\PS}$ is the set of all interpretations over $\PS$.
An interpretation $\omega$ is a model of a formula $\phi$ if and
only if the assignment of the variables of $\PS$ in $\phi$
according to $\omega$ leads $\phi$ to be evaluated to
$1$. $\phi(\omega)$ denotes the truth value given to 
$\phi$ by $\omega$.
The languages we consider in the following are described in
Darwiche and Marquis' compilation map~\cite{compilationMap}. 
They are subsets of the following \nnf\ language (composed of circuits, alias DAG-shaped
"formulae")~\cite{darwiche1999compiling,darwiche2001decomposable}.

\begin{definition}[\nnf]~
  \nnf\ is the set of rooted, directed acyclic graphs (DAGs) where
  each leaf node is labeled with $\bot$, $\top$, $x$ or $\neg x$, $x \in
  \PS$; and each internal node is labeled with $\land$ or $\lor$ and
  can have arbitrarily many children.
\end{definition}

Figure~\ref{fig:nnf} is a \nnf\ formula which represents some parts of the constraints used in our running
example. This formula is equivalent to the \cnf\ formula 
$(\lnot A_{1} \lor B)\land (\lnot A_{1} \lor \lnot A_{2}) \land (B \lor \lnot A_{2}) \land (\lnot A_{2} \lor C)$, 
which is also a \nnf\ formula, since \cnf\ is a subset of \nnf.
\begin{figure}
  \centering
  \subfigure[\label{fig:nnf}A \cnf\ formula]{
    \begin{tikzpicture}[thick,scale=0.8, every node/.style={scale=0.8}]
      \tikzstyle{nnfn}=[]
      \node[nnfn] (AND) at (0,0) {$\land$};
      \node[nnfn] (OR1) at (-1.5,-1) {$\lor$};
      \node[nnfn] (OR2) at (-.5,-1) {$\lor$};
      \node[nnfn] (OR3) at (.5,-1) {$\lor$};
      \node[nnfn] (OR4) at (1.5,-1) {$\lor$};
      \node[nnfn] (NA1) at (-1.5,-2) {$\neg A_1$};
      \node[nnfn] (B) at (-.5,-2) {$B$};
      \node[nnfn] (NA2) at (.5,-2) {$\neg A_2$};
      \node[nnfn] (C) at (1.5,-2) {$C$};
      \draw (NA1) -- (OR1) -- (B);
      \draw (NA1) -- (OR2) -- (NA2);
      \draw (B) -- (OR3) -- (NA2);
      \draw (NA2) -- (OR4) -- (C);
      \draw (OR1) -- (AND) -- (OR2);
      \draw (OR3) -- (AND) -- (OR4);
    \end{tikzpicture}
  }
  ~ 
  \subfigure[\label{fig:dnnf}A \Dnnf\ (\dnf) formula]{
    \begin{tikzpicture}[thick,scale=0.8, every node/.style={scale=0.8}]
      \tikzstyle{nnfn}=[]
      \node[nnfn] (OR) at (0,0) {$\lor$};
      \node[nnfn] (AND1) at (-1,-1) {$\land$};
      \node[nnfn] (AND2) at (0,-1) {$\land$};
      \node[nnfn] (AND3) at (1.5,-1) {$\land$};
      \node[nnfn] (A1) at (-2,-2) {$A_1$};
      \node[nnfn] (NA2) at (-1,-2) {$\neg A_2$};
      \node[nnfn] (B) at (0,-2) {$B$};
      \node[nnfn] (NA1) at (1,-2) {$\neg A_1$};
      \node[nnfn] (A2) at (2,-2) {$A_2$};
      \node[nnfn] (C) at (3,-2) {$C$};
      \draw (A1) -- (AND1) -- (NA2) -- (AND1) -- (B);
      \draw (NA2) -- (AND2) -- (NA1);
      \draw (B) -- (AND3) -- (NA1) -- (AND3) -- (A2) -- (AND3) -- (C);
      \draw (AND1) -- (OR) -- (AND2) -- (OR) -- (AND3);
    \end{tikzpicture}
  }
  \caption{\nnf\ formulae}
  \vspace{-0.5cm}
\end{figure}
The size of a formula $\phi$ in \nnf, denoted $|\phi|$, is the number of  
arcs in it. 
For any node $N$ of a \nnf\ formula $\phi$, 
$\mathit{Vars}(N)$ denotes the set of variables labeling the leaf nodes
which can be reached from $N$; $\mathit{Vars}(\phi)$ is equal to $\mathit{Vars}(N)$
 where $N$ is the root node of $\phi$.

In ~\cite{compilationMap}, several properties have been 
considered on the \nnf\ language, leading to a
family of  languages which are subsets of \nnf. We have already 
noted that \cnf\ is one of these subsets, but many more languages have been
considered. In this article, each language $L$ from  
the knowledge compilation map, namely \nnf, \Dnnf,
\dnnf, \snnf, \fnnf, \dDnnf, \sdDnnf, \bdd, \fbdd, \obdd,
\oobdd, \dnf, \cnf, \pi, \ip\ and \mods, is considered. We add to them the languages
\tDnnf\ and \sDnnf\ which have been introduced later \cite{PipatsrisawatDarwiche08}.
For space reasons, we recall here only some of the properties
and the corresponding languages. More details are to be found in
~\cite{compilationMap,PipatsrisawatDarwiche08}.
\begin{definition}[decomposability]~\cite{darwiche1999compiling,darwiche2001decomposable}
  A \nnf\ formula $\phi$ satisfies the {\em decomposability property} if for each
  conjunction $C$ in $\phi$, the conjuncts of $C$ do not share
  variables. That is, if $C_1,...,C_n$ are the children of an and-node
  $C$, then $\mathit{Vars}(C_i) \cap \mathit{Vars}(C_j) = \emptyset$ for $i \neq j$.
\end{definition}
The subset of \nnf\ which satisfies the decomposability property is called \Dnnf. Note
that the \dnf\ language is a subset of \Dnnf\ (assuming without loss of generality that 
all literals of the terms refer to different variables). 

A number of queries and transformations have been considered in the knowledge compilation map.
A query corresponds to a computational problem which consists in extracting some
information from a given \nnf\ formula $\phi$, without modifying it. A transformation aims at generating
an \nnf\ formula. Queries and transformations are also viewed as properties which are satisfied or not by a given subset $L$ of \nnf:
$L$ is said to satisfy a given query/transformation precisely when there exists a polynomial-time
algorithm for answering the query/achieving the transformation provided that the input is a formula from $L$.
Among others, the following queries and transformations have been considered in the knowledge 
compilation map:
\begin{itemize}
\item \co\ (consistency): a language $L$ satisfies the consistency query \co\ if and only
  if there exists a polynomial-time algorithm that maps every formula
  $\phi$ from $L$  to $1$ if $\phi$ is consistent (i.e.,
  $\phi$ has at least one model), and to $0$ otherwise;
\item $\cd$ (conditioning) : a language $L$ satisfies the $\cd$ transformation if and
  only if there exists a polynomial-time algorithm that maps every
  formula $\phi$ from $L$ and a consistent term $\gamma$ to a formula of $L$ that is logically
  equivalent to $\phi \mid \gamma$.
\end{itemize}




In the following, two aggregation functions are considered: the standard summation $\Sigma$
aggregator leading to an utilitarist aggregation of values as well as $\mathit{leximax}$, which is
a refinement of $\mathit{max}$ and leads to an egalitarist aggregation of values \cite{moulin1991axioms}; when $\mathit{leximax}$
is considered,  a solution $\omega$ is considered at least as preferred as a solution $\omega'$ 
when $f(\omega) \leq f(\omega')$, where $f(\omega)$ (resp. $f(\omega')$) is the $n$-vector of scores 
associated with $\omega$ (resp. $\omega'$) and reordered in a decreasing way; $\leq$ denotes here
the lexicographic ordering over the vectors of scores. Thus, one prefers to minimize first the penalties
stemming from the most important criteria (the ones of highest weights), then those stemming from the second most important criteria, and so on. 
For instance, provided that $\PS = \{A_1, A_2, B_1, C_1\}$, given the weighted base 
$\{(A_2 \wedge C_1, 2), (B_1 \wedge \neg C_1, 1)\}$,
$\omega = \{(A_1, 1), (A_2, 0), (B_1, 1), (C_1, 0)\}$ is (strictly) preferred to
$\omega' = \{(A_1, 0), (A_2, 1), (B_1, 1), (C_1, 1)\}$ because
$f(\omega) = (1, 0) < f(\omega') = (2, 0)$.
Unlike $\Sigma$, no balance between criteria is possible when $\mathit{leximax}$ is used.

In the following we consider that the pseudo-Boolean optimization functions 
are represented as weighted bases.
$\mathcal{G}$ denotes the set of  ("general") representations, i.e., when the $\phi_i$ are any \nnf\ representations.
One also considers several restrictions on the $\phi_i$ (leading to some subsets of $\mathcal{G}$) which prove of interest:
\begin{itemize}
\item {\em linear representations} are when each $\phi_i$ is a literal; $\mathcal{L}$ is the corresponding language.
\item {\em quadratic representations} are when each $\phi_i$ is a term of size at most 2; $\mathcal{Q}$ is the corresponding language.
\item {\em polynomial representations} are when each $\phi_i$ is a term; $\mathcal{P}$ is the corresponding language.
\end{itemize}

We have the obvious inclusions:
$$\mathcal{L} \subset \mathcal{Q} \subset \mathcal{P} \subset \mathcal{G}.$$

$\mathcal{Q}$ is of interest for evaluating the complexity of the optimization problem for $\mathcal{P}$ because 
Rosenberg proved that every polynomial representation of a pseudo-Boolean function $f$ can be associated
in polynomial time with a quadratic representation of $f$, without
altering the set of optimal solutions~\cite{rosenberg1975reduction}. 

For each language $\mathcal{F}$ among $\mathcal{L}$, $\mathcal{Q}$, $\mathcal{P}$, $\mathcal{G}$, we also consider
the subset $\mathcal{F}^+$ of $\mathcal{F}$ obtained by assuming that the only literals occurring in any $\phi_i$
are positive ones. Finally, for each of the resulting languages $\mathcal{F}$, we consider the subset $\mathcal{F}_+$
of it obtained by assuming that each weight $w_i$ is from $\mathbb{R}^+$. 
Thus, for instance, $\mathcal{P}^+$ denotes the set of all polynomial representations based on
 positive literals (i.e., the $\phi_i$ are positive terms), $\mathcal{Q}_+$ denotes the set of all quadratic representations
 with non-negative weights, and $\mathcal{L}_+^+$ denotes the set of 	all linear representations, based on positive
 literals and with non-negative weights.

The optimization query (\opt) is defined as follows:

\begin{definition}[\opt]~\\
\vspace{-4mm}
\begin{itemize}
\item \opt\ (optimization):  a language $L$ satisfies \opt\ given $\mathcal{F}$ and $\oplus$ if
  and only if there exists a polynomial-time algorithm that maps every
  formula $\phi$ from $L$ and every representation from $\mathcal{F}$
  of a pseudo-Boolean optimization function $f$ to an optimal solution
  of $\phi$ given $f$ and $\oplus$ when a feasible solution exists, and to "no
  solution" otherwise.
  \end{itemize}
\end{definition}

To make things more precise, we note \opt[$L$,
$\mathcal{F}$, $\oplus$]  the optimization problem for $L$ when the weighted base
representing $f$ is in $\mathcal{F}$ and $\oplus$ is the aggregator.

It is important to understand that the complexity of the optimization
query for $L$ depends on the {\em representation} of the pseudo-Boolean
optimization function $f$ (this representation is part of the input),
and not on the function $f$ itself. Indeed, on the one hand, consider
any pseudo-Boolean optimization function $f$ which is represented in
an explicit way (i.e., as the weighted base $\{(\omega, f(\omega)) \mid \omega \in
\Omega_{\PS}\}$).  Any $L$ in \nnf\ satisfies \opt\ in such a case because $\Omega_{\PS}$ is part of the input: 
just consider each $\omega
\in \Omega_{\PS}$ and check in polynomial time whether $\omega$ is a
model of $\phi$. If so, compare $f(\omega)$ 
with the optimal value $\mathit{opt}$ obtained so far and replace $\mathit{opt}$ by
$f(\omega)$ if $f(\omega)$ is better than $\mathit{opt}$ (and in this case store
$\omega$).
On the other hand, consider the case of a constant function
$f$, represented by an empty weighted base; in this case, solving the optimization
problem for $L$ amounts to determining whether $L$ satisfies or not
\co, which is not doable in polynomial time for many subsets of
\nnf. 

In the following , we analyze the computational complexity of \opt\ for the languages $L$
from the knowledge compilation map satisfying \co, namely \Dnnf,
\dDnnf, \fbdd, \oobdd, \dnf, \ip, \pi\ and \mods, together with
\tDnnf. We ignore \obdd\ (resp. \sDnnf) because only one formula is
considered for \opt, which prevents from ordering (resp. vtree) clashes;
thus the results will be exactly the same as the ones for \oobdd\
(resp. \tDnnf). As to the representation of weighted bases, one
considers the languages $\mathcal{L}$, $\mathcal{Q}$, $\mathcal{P}$, $\mathcal{G}$, 
and their restrictions to positive literals and/or positive weights. Finally,
we consider both $\Sigma$ and $\mathit{leximax}$ as aggregators.

The rationale for rejecting languages $L$ not satisfying \co\ is obvious: if determining
whether a feasible solution exists is {\sf NP}-hard, then determining whether an optimal one exists is 
{\sf NP}-hard as well:

\begin{proposition}\label{prop:obvious}
  If $L$ does not satisfy \co\ unless {\sf P = NP}, then \opt[$L$,
  $\mathcal{F}$, $\oplus$] is {\sf NP}-hard whatever $\mathcal{F}$ and
  $\oplus$.\footnote{Proofs of propositions are located at the end of
    the paper.}
\end{proposition}

Specifically, this is the case for all languages $L$ which do not quality as target languages for knowledge compilation, like
\cnf\ and \nnf\ \cite{compilationMap}.

\section{Linear Representations}

We first consider the case of linear representations of 
pseudo-Boolean objective functions.
A pseudo-Boolean objective function $f$ is said to be linear 
when it has a linear representation. Obviously enough, this is not the case
for every pseudo-Boolean objective function. Especially, when $\oplus = \Sigma$ is the aggregator,
only modular functions can be represented linearly.
Nevertheless, modular functions $f$ are enough for many problems.
Thus, a long list of scenarios where such functions are used is reported  
in ~\cite{boros2002pseudo}.
Clearly, \opt[$L$, $\mathcal{L}$, $\Sigma$] is {\sf NP}-hard 
for many $L$, like in the case of pseudo-Boolean optimization \cite{PBO} (i.e., when 
$L$ consists of conjunctions of equations or inequations over Boolean variables) or more generally
in the case of mixed integer programming \cite{MILP}.

%
%
%
%
%

\subsection{Intractable Cases}

Let us start with the subsets of \nnf\ for which \opt\ given $\mathcal{L}$
is intractable. We have already shown that if $L$ does not satisfy \co\ unless {\sf P = NP}, then 
\opt[$L$, $\mathcal{F}$, $\oplus$]  is {\sf NP}-hard whatever $\mathcal{F}$ and $\oplus$.
Obviously, this hardness result still holds when $\mathcal{F} = \mathcal{L}$.
The following proposition shows that the converse implication does not hold:
it can be the case that $L$ satisfies \co, and that \opt[$L$, $\mathcal{L}$, $\oplus$]  is {\sf NP}-hard:

\begin{proposition}\label{prop:pi:np}
\opt[\pi, $\mathcal{L}^+_+$, $\oplus$]  is {\sf NP}-hard for $\oplus = \Sigma$ and $\oplus = \mathit{leximax}$.
\end{proposition}

As a consequence, \opt[\pi, $\mathcal{F}$, $\oplus$]  is {\sf NP}-hard (with $\oplus = \Sigma$ or $\oplus = \mathit{leximax}$)
for every superset $\mathcal{F}$ of $\mathcal{L}^+_+$.


%


\subsection{Tractable Cases}

We now define the concepts of partial interpretation and model generator
which will be used in the following lemmas.
\begin{definition}[partial interpretation] 
  Let $V$ be a set of propositional variables ($V \subseteq \text{PS}$). A {\em partial
  interpretation} $\omega_V$ over $\text{PS}$ is a set of pairs $(v \in
  V,\{0,1\})$ where each $v \in V$ is the first projection of exactly one
  pair in $V$. It is also viewed as the term $\bigwedge_{(v, 0) \in \omega_V} \neg v
  \wedge \bigwedge_{(v, 1) \in \omega_V} v$.
\end{definition}
The term \textit{partial} comes from the fact that $\omega_V$ may be
extended by adding the missing pairs $(v \in PS\setminus V,
\{0,1\})$ to get a ``full'' interpretation over $\text{PS}$.
  Let $\omega$ be an interpretation on $\text{PS}$ and $\omega_V$ be a
  partial interpretation on $\text{PS}$. If $\omega_V \subseteq \omega$, then 
  $\omega$ is an {\em extension} of $\omega_V$.
When all the extensions generated from a partial interpretation
$\omega_V$ satisfy a formula $\phi$, and no other interpretation satisfies $\phi$,
$\omega_V$ is said to be a {\em model generator} of $\phi$.
\begin{definition}[model generator]~
  Let $\omega_V$ be a partial interpretation such that the set of its
  extensions is equal to the set of models of $\phi$. $\omega_V$
  is a {\em model generator} of $\phi$.
\end{definition}

We now describe a polynomial-time algorithm for computing an optimal
model of a \Dnnf\ formula $\phi$ given a linear representation of a pseudo-Boolean function. 
This algorithm returns "no solution" when $\phi$ is inconsistent. Otherwise, it generates an optimal 
solution in a bottom-up way, from the leaves to the root of the \Dnnf\ formula.
The correctness of the optimization algorithm is based on a number of lemmas.

\begin{lemma}\label{lemma:optFromModGen}
  Let $\omega_V$ be a model generator of an \nnf\ formula $\phi$.
  Given a linear representation of a pseudo-Boolean function $f$, one
  can generate in polynomial time an optimal model $\omega^*$ of $\phi$ given
  $f$ such that $\omega^*$ extends $\omega_V$.
\end{lemma}


Furthermore, given a \nnf\ formula $\phi$ which reduces to a leaf, 
computing a model generator of $\phi$ when it exists
and determining that no such generator exists otherwise is easy:

\begin{lemma}\label{lemma:leafModGen}
  Let $\phi$ be a \nnf\ formula such that $|\phi|=1$. 
  One can compute in constant time a model generator of $\phi$ when it exists
  and determine that no such generator exists otherwise.
\end{lemma}


This addresses the base cases. It remains to consider the general case.
Let us first focus on decomposable \AN nodes. The decomposability property 
implies that the children of these nodes do not share variables, thus the model generator of a \AN node can be derived from the union of the model generators of its children.

\begin{lemma}\label{lemma:andModGen}
  Let $\phi = \bigwedge_{i=1}^k \phi_i$ be a \nnf\ formula rooted at a
  decomposable \AN node. Suppose that each $\phi_i$ ($i \in \{1, \ldots, k\}$) is
  associated with its model generator $\omega_i$ when such a generator exists. Then
  a model generator of $\phi$ can be computed in linear time from $\omega_1, \ldots, \omega_k$,
  when it exists.
\end{lemma}


The last kind of \Dnnf\ nodes to be considered are \ON nodes. \ON
nodes are the nodes. We highlight here that the model generator of an 
\ON node can be derived from the ones of its children by selecting a model 
generator leading to the best optimal value.
Note that in this case, the model generator
does not represent all the models of the formula rooted at the \ON
node, but a set of models which contain at least one optimal model.

\begin{lemma}\label{lemma:orModGen}
  Let $\phi = \bigvee_{i=1}^k \phi_i$ be a NNF formula rooted at an \ON node.
  Suppose that each $\phi_i$ ($i \in \{1, \ldots, k\}$) is
  associated with its model generator $\omega_i$ when such a generator exists. Then, when it exists,
  for any linear representation of a pseudo-Boolean function $f$, one can select in linear time 
  among $\omega_1, \ldots, \omega_k$ a partial interpretation $\omega_{\mathit{opt}}$ ($\mathit{opt} \in \{1, \ldots, k\}$) which is extended 
  by an optimal model of $\phi$ given $f$.
\end{lemma}



Taking advantage of the previous lemmas, one gets immediately a
polynomial-time algorithm for generating an optimal model of a \Dnnf\
formula $\phi$ (when it exists) given a linear representation of a pseudo-Boolean
function $f$.
%
%
As a consequence, we get that \Dnnf\ satisfies \opt\ given $\mathcal{L}$:

\begin{proposition}\label{prop:dnnf:p}
\opt[\Dnnf, $\mathcal{L}$, $\oplus$] is in {\sf P} for $\oplus = \Sigma$
 and $\oplus = \mathit{leximax}$.
\end{proposition}


Consequently, optimization also is tractable for each subset of \Dnnf,
including  \dnf, \ip, \dDnnf, \fbdd, \obdd, \mods\ and \tDnnf.
In the case $\oplus = \Sigma$, this proposition coheres with a result reported in \cite{Kimmigetal12}
which shows how to solve in polynomial time the weighted model
counting problem when the input is a smooth \Dnnf\ formula (that is,
for each disjunction node of the \Dnnf\ formula, its children mention
the same variables). Indeed, it turns out that $(\mathbb{R},
\mathit{min}, +, +\infty, 0)$ is a commutative semiring and that, in
this semiring, the weighted model count associated with a \Dnnf\
formula $\phi$ given the weights $\{(l_i, w_i) \mid l_i$ literal over
$\PS$$\}$ is, under some computationally harmless
conditions,\footnote{$\phi$ must be consistent, smooth and every
  variable of $\PS$ must occur in it; the consistency condition can be
  decided in linear time when $\phi$ is a \Dnnf\ formula; furthermore,
  every \Dnnf\ formula can be be turned in time linear in it and in
  the number of variables of $\PS$ into a smoothed \Dnnf\ formula in
  which every variable of $\PS$ occurs.}  equal to the value of an
optimal model of $\phi$ for the pseudo-Boolean function represented by
the weighted base $\{(l_i, w_i) \mid l_i$ literal over $\PS$$\}$.  In
our case, one not only computes such a value, but also returns an
optimal solution $\omega^*$ leading to this value. Furthermore,
our tractability result also applies to the case $\oplus = \mathit{leximax}$.
Finally, it is important to note that this tractability result cannot be extended
to the full family of $\mathit{OWA}_W$ aggregators
 \cite{Yager88} (this explains why we focused
on specific $\mathit{OWA}_W$ aggregators, namely $\Sigma$ and $\mathit{leximax}$):

\begin{proposition}\label{prop:owa}
Let $L$ be any subset of \nnf\ containing $\top$. For some $W$, 
 \opt[$L$,$\mathcal{L}_+$,$\mathit{OWA}_W$] is {\sf NP}-hard.
\end{proposition}

This result is based on a polynomial-time reduction from \opt[$L$,$\mathcal{Q_+}$,$\Sigma$] to
\opt[$L$,$\mathcal{L}_+$,$\mathit{OWA}_W$] for some $W$ and uses the Proposition \ref{prop:nnf:q+0:np} found in the next section.

\section{Non-Linear Representations}

Optimization processes are typically considered for making a choice between multiple
solutions. In some cases, linear pseudo-Boolean functions are not expressive
enough to encode the preference relations of interest.  
Interestingly, many non-linear optimization functions $f$ admit
linearizations, i.e., one can associate in polynomial time with an
optimization problem based on such an $f$ an optimization problem
based on a linear function which has ``essentially'' the same optimal
solutions as the original problem.  Such a linearization process is
achieved by adding new variables and constraints to the problem; those
variables and constraints depend on $f$, and ``essentially" means here
any optimal solution of the linearized problem must be projected onto
the original variables to lead to a solution of the initial optimization
problem. Such an approach is used in practice in the
non-linear track of the pseudo-Boolean evaluation \cite{PBO}.
For example, a linearization process can be applied to polynomial
representations of pseudo-Boolean optimizations functions (see below)
by adding a constraint and an integer variable (decomposed using
Boolean variables) for each term. A number of optimization procedures
have also been proposed for some aggregations of linear
representations of pseudo-Boolean functions, for instance when the aggregator
under consideration belongs to the $\mathit{OWA}_W$ family \cite{ogryczak2003solving,Bolandetal06,GalandSpanjaard12};
in the corresponding encodings the number of constraints and
integer variables to be added are typically quadratic in the number of functions to be 
aggregated.

Unfortunately, such linearization techniques are not convenient in the general case when
the set of feasible solutions is represented by a \Dnnf\
formula. Indeed, if one wants to exploit the optimization algorithm
presented in the previous section, then it is necessary to compute a \Dnnf\
formula equivalent to the conjunction of the \Dnnf\ formula coming
from the initial optimization problem with the new constraints
issued from the linearization process. The point is that such a
computation cannot be achieved in polynomial time in the general case
(\Dnnf\ does not satisfy the conjunctive closure
transformation~\cite{darwiche1999compiling,darwiche2001decomposable}).
The complexity results we present in the following implies that this
is actually the case for many non-linear functions, unless {\sf P =
  NP}.  Accordingly, the benefits offered by the compilation-based
approach (i.e., solving efficiently some optimization queries when $f$
varies) are lost in the general case when non-linear representations are considered.

%

\subsection{Polynomial Representations}

Let us start with the polynomial representations of pseudo-Boolean objective functions.
Unlike the linear case, every pseudo-Boolean objective function $f : \mathbb{R}^n \rightarrow \mathbb{R}$ has a polynomial representation.
Indeed, $\{(\omega, f(\omega)) \mid \omega \in \Omega_{\PS}\}$ is a polynomial representation
of $f$. 


\opt\ turns out to be {\sf NP}-hard for almost all valuable subsets of \nnf,
even under some strong conditions on the representation of the optimization function:

\begin{proposition}\label{prop:nnf:q+0:np}
For each subset $L$ of \nnf\ containing $\top$ (which is the case for all the subsets of \nnf\ considered
in the paper, including \mods), \opt[$L$, $\mathcal{Q}^+$, $\oplus$] and \opt[$L$, $\mathcal{Q}_+$, $\oplus$] are {\sf NP}-hard
with $\oplus = \Sigma$ or $\oplus = \mathit{leximax}$, even under the restriction $\phi = \top$.
\end{proposition}

\renewcommand\arraystretch{1.4277}
\newcommand\summaryScale{.85}
\begin{figure*}[b]
  \centering
\scalebox{\summaryScale}{
\begin{tabular}{|c||c|c|c|c||c|c|c|c||c|c|c|c|c|c|}\hline
~       & $\mathcal{G}$                              & $\mathcal{G}^+$                            & $\mathcal{G}_+$                            & $\mathcal{G}^+_+$                               & $\{\mathcal{P},\mathcal{Q}\}$               & $\{\mathcal{P},\mathcal{Q}\}^+$            & $\{\mathcal{P},\mathcal{Q}\}_+$             & $\{\mathcal{P},\mathcal{Q}\}^+_+$& $\mathcal{L}$                                & $\mathcal{L}^+$                              & $\mathcal{L}_+$                              & $\mathcal{L}^+_+$\\
\hline \hline
\Dnnf\  & \n$^{p\ref{prop:nnf:q+0:np}}_{p\ref{prop:l:g+0:nfpt}}$ & \n$^{p\ref{prop:nnf:q+0:np}}_{p\ref{prop:l:g+0:nfpt}}$ & \n$^{p\ref{prop:nnf:q+0:np}}_{p\ref{prop:l:g+0:nfpt}}$ & \n$^{p\ref{prop:obddpi:np}}_{p\ref{corol:obddpi:g++:nfpt}}$ & \fpt$^{p\ref{prop:nnf:q+0:np}}_{p\ref{prop:cdco:p:p}}$ & \fpt$^{p\ref{prop:nnf:q+0:np}}_{p\ref{prop:cdco:p:p}}$ & \fpt$^{p\ref{prop:nnf:q+0:np}}_{p\ref{prop:cdco:p:p}}$ & \fpt$^{p\ref{prop:obddpi:np}}_{p\ref{prop:cdco:p:p}}$& \y$^{p\ref{prop:dnnf:p}}$                         & \y$^{p\ref{prop:dnnf:p}}$                         & \y$^{p\ref{prop:dnnf:p}}$                         & \y$^{p\ref{prop:dnnf:p}}$\\
\hline
$(*)$ \tDnnf,\oobdd\ & \n$^{p\ref{prop:nnf:q+0:np}}_{p\ref{prop:l:g+0:nfpt}}$ & \n$^{p\ref{prop:nnf:q+0:np}}_{p\ref{prop:l:g+0:nfpt}}$ & \n$^{p\ref{prop:nnf:q+0:np}}_{p\ref{prop:l:g+0:nfpt}}$ & \n$^{p\ref{prop:obddpi:np}}_{p\ref{corol:obddpi:g++:nfpt}}$ & \fpt$^{p\ref{prop:nnf:q+0:np}}_{p\ref{prop:cdco:p:p}}$ & \fpt$^{p\ref{prop:nnf:q+0:np}}_{p\ref{prop:cdco:p:p}}$ & \fpt$^{p\ref{prop:nnf:q+0:np}}_{p\ref{prop:cdco:p:p}}$ & \fpt$^{p\ref{prop:obddpi:np}}_{p\ref{prop:cdco:p:p}}$ & \y$^{p\ref{prop:dnnf:p}}$                         & \y$^{p\ref{prop:dnnf:p}}$                         & \y$^{p\ref{prop:dnnf:p}}$                         & \y$^{p\ref{prop:dnnf:p}}$\\
\hline
\dnf, \ip, \mods\   & \n$^{p\ref{prop:nnf:q+0:np}}_{p\ref{prop:l:g+0:nfpt}}$ & \n$^{p\ref{prop:nnf:q+0:np}}_{p\ref{prop:l:g+0:nfpt}}$ & \n$^{p\ref{prop:nnf:q+0:np}}_{p\ref{prop:l:g+0:nfpt}}$ & \y$^{p\ref{prop:dnf:p}}$& \fpt$^{p\ref{prop:nnf:q+0:np}}_{p\ref{prop:cdco:p:p}}$ & \fpt$^{p\ref{prop:nnf:q+0:np}}_{p\ref{prop:cdco:p:p}}$ & \fpt$^{p\ref{prop:nnf:q+0:np}}_{p\ref{prop:cdco:p:p}}$ & \y$^{p\ref{prop:dnf:p}}$ & \y$^{p\ref{prop:dnnf:p}}$                         & \y$^{p\ref{prop:dnnf:p}}$                         & \y$^{p\ref{prop:dnnf:p}}$                         & \y$^{p\ref{prop:dnnf:p}}$                            \\
\hline
\pi\    & \n$^{p\ref{prop:pi:np}}_{p\ref{prop:l:g+0:nfpt}}$     & \n$^{p\ref{prop:pi:np}}_{p\ref{prop:l:g+0:nfpt}}$     & \n$^{p\ref{prop:pi:np}}_{p\ref{prop:l:g+0:nfpt}}$    & \n$^{p\ref{prop:pi:np}}_{p\ref{corol:obddpi:g++:nfpt}}$   & \fpt$^{p\ref{prop:pi:np}}_{p\ref{prop:cdco:p:p}}$     & \fpt$^{p\ref{prop:pi:np}}_{p\ref{prop:cdco:p:p}}$    & \fpt$^{p\ref{prop:pi:np}}_{p\ref{prop:cdco:p:p}}$     & \fpt$^{p\ref{prop:pi:np}}_{p\ref{prop:cdco:p:p}}$  & \fpt$^{p\ref{prop:pi:np}}_{p\ref{prop:cdco:p:p}}$ & \fpt$^{p\ref{prop:pi:np}}_{p\ref{prop:cdco:p:p}}$ & \fpt$^{p\ref{prop:pi:np}}_{p\ref{prop:cdco:p:p}}$ & \fpt$^{p\ref{prop:pi:np}}_{p\ref{prop:cdco:p:p}}$ \\
\hline
\end{tabular}
}

\caption{\label{fig:summary}Complexity of \opt\ for subsets of \dnnf,
  when $\oplus = \Sigma$ or $\oplus = \mathit{leximax}$. \y\ means
  ``satisfies'', \fpt\ means ``does not satisfy unless {\sf P = NP}
  but is in {\sf FPT} with parameter $n$'' and \n\ means ``does not
  satisfy unless {\sf P = NP} and is {\sf NP}-hard as soon as $n \geq
  2$". 
  In each cell, the exponent (resp. subscript) indicates the
  proposition from which the result reported in the cell comes when
  $n$ is unbounded (resp. when $n$ is bounded).  $(*)$ \opt[\oobdd,
  $\mathcal{G}$, $\oplus$] (resp. \opt[\tDnnf, $\mathcal{G}$,
  $\oplus$]) is in {\sf FPT} with parameter $n$ for $\oplus = \Sigma$
  and $\oplus = \mathit{leximax}$ when each $\phi_i$ is in \oobdd\
  (resp. \tDnnf).}
\end{figure*}

\subsection{General Representations}

Clearly enough, as a consequence of the results reported in the previous section, the optimization problem \opt[$L$, $\mathcal{G}$, $\oplus$] is 
intractable in the general case. One way to recover tractability consists in imposing some strong restrictions on both the language $L$ and the weighted base:

\begin{proposition}\label{prop:dnf:p}
\opt[\dnf, $\mathcal{G}^+_+$, $\oplus$] is in {\sf P} for $\oplus = \Sigma$ or $\oplus = \mathit{leximax}$.
\end{proposition}


Both restrictions are needed; especially, the result cannot be generalized to the other subsets of \Dnnf\
considered in this paper (except of course \ip\ and \mods\ which are subsets of \dnf):

\begin{proposition}\label{prop:obddpi:np}
\opt[\oobdd, $\mathcal{Q}^+_+$, $\oplus$] and \opt[\pi, $\mathcal{Q}^+_+$, $\oplus$] are {\sf NP}-hard
when $\oplus = \Sigma$ or $\oplus = \mathit{leximax}$. 
\end{proposition}

\section{Some Fixed-Parameter Tractability Results}




When considering representations of pseudo-Boolean functions based on weighted formulae, a natural
restriction is to bound the cardinality $n$ of the weighted base, since this amounts to considering only
a restricted number of criteria of interest. This corresponds to a case for which the user's preferences are, so to say, 
``simple" ones. It can be expected that the complexity of \opt\ increases as
a polynomial in $n$. In the following we show that this is the case for the family $\mathcal{P}$ of
polynomial representations (under some harmless conditions on $L$):


\begin{proposition}\label{prop:cdco:p:p}
  Let $L$ be a propositional language that satisfies \cd\ and
  \co. \opt[$L$, $\mathcal{P}$, $\oplus$] is in {\sf FPT} with parameter $n$
  for $\oplus = \Sigma$ and $\oplus = \mathit{leximax}$.
\end{proposition}


Contrastingly, as soon as we consider general representations of the objective functions, strong assumptions are necessary to get positive results, 
even when $n$ is bounded:

\begin{proposition}\label{prop:obdd:obdd:fpt}
Provided that each $\phi_i$ used in the representation of the weighted base is an \oobdd\ representation (resp. a \tDnnf\ representation), 
\opt[\oobdd, $\mathcal{G}$, $\oplus$] (resp. \opt[\tDnnf, $\mathcal{G}$, $\oplus$]) is in {\sf FPT} with parameter $n$  for $\oplus = \Sigma$ and $\oplus = \mathit{leximax}$.
\end{proposition}


Unfortunately, when considering the representation language
$\mathcal{G}$ in the general case, the results are again mostly
negative.
While Proposition \ref{prop:dnf:p} states that 
when weighted bases are represented in $\mathcal{G}^+_+$, \dnf\ admits a
polynomial-time optimization algorithm, it turns out that relaxing any of the
two conditions imposed on $\mathcal{G}$ implies that there is no such
algorithm for any of the \nnf\ languages we consider (unless {\sf P = NP}), even when $n$ is bounded:

\begin{proposition}\label{prop:l:g+0:nfpt}
For each subset $L$ of \nnf\ containing $\top$, \opt[$L$,$\mathcal{G}^+$, $\oplus$] 
and \opt[$L$,$\mathcal{G}_+$, $\oplus$] are {\sf NP}-hard under the restriction $n \geq 2$ and $\phi = \top$
 for $\oplus = \Sigma$ and $\oplus = \mathit{leximax}$.
\end{proposition}

In particular, this proposition shows that \opt[\dnf, $\mathcal{G}^+$, $\oplus$] and 
\opt[\dnf, $\mathcal{G}_+$, $\oplus$] are {\sf NP}-hard, even when $n \geq 2$,
 for $\oplus = \Sigma$ and $\oplus = \mathit{leximax}$. Hence switching
 from $\mathcal{G}_+^+$ to any of its supersets $\mathcal{G}_+$ or $\mathcal{G}^+$
 has a strong impact on complexity (compare Proposition \ref{prop:l:g+0:nfpt}
 with Proposition \ref{prop:dnf:p}).





Finally, we derived the following proposition showing
that no result similar to Proposition \ref{prop:dnf:p}
holds for some other interesting subsets of \Dnnf\ which are not subsets of \dnf\
(namely \oobdd\ and \pi), even when the cardinality of the weighted base is supposed to be
bounded:

\begin{proposition}\label{corol:obddpi:g++:nfpt}
\opt[\oobdd,$\mathcal{G}^+_+$, $\oplus$] and \opt[\pi,$\mathcal{G}^+_+$, $\oplus$] 
are {\sf NP}-hard under the restriction $n \geq 2$
for $\oplus = \Sigma$ and $\oplus = \mathit{leximax}$.
\end{proposition}

\section{Conclusion}
In this article, we investigated the feasibility of a compilation-based approach to optimization, 
where the set of admissible solutions is represented by a hard constraint (a propositional formula) $\phi$
which is compiled during an off-line phase, and a set of representations of pseudo-Boolean objective functions $f$ 
available only at the on-line phase. Two aggregators have been considered
($\Sigma$ and $\mathit{leximax}$). Our main results are summarized in Fig.\ref{fig:summary}.

%


Our study shows that the optimization query remains intractable in
most cases except for linear representations of the objective
function. In this case, it makes sense to compile the hard constraint
$\phi$ into a \Dnnf\ representation since \Dnnf\ is the more succinct
language among those considered here, offering tractable optimization.
However, we have shown that when forcing the weights and the literals
of the objective function to be positive, the optimization problem for
\dnf\ becomes tractable. We have also investigated the case when the
number of weighted formulae in the representation of the objective
function is bounded. We found out that under this hypothesis, the
optimization query becomes tractable for the polynomial
representations of the objective functions. Finally, it is worth
noting that while the languages {\tt ADD} \cite{BaharFGHMPS93}, {\tt
  SLDD} \cite{Wilson05} and {\tt AADD} \cite{SannerM05} of valued
decision diagrams can be used for representing and handling
pseudo-Boolean objective functions $f$, they are not suited to our
compilation-based approach to optimization.
Indeed, the compilation of $f$ in any of those
languages cannot be achieved in polynomial time in the general case;
each time a new objective function $f$ is considered, a
(time-consuming) compilation phase must be undertaken.




\vfill
\pagebreak
\bibliography{aaai}

\begin{thebibliography}{}

\bibitem[\protect\citeauthoryear{Argelich \bgroup et al\mbox.\egroup
  }{2010}]{p2cudf}
Argelich, J.; Berre, D.~L.; Lynce, I.; Silva, J. P.~M.; and Rapicault, P.
\newblock 2010.
\newblock Solving linux upgradeability problems using boolean optimization.
\newblock In Lynce, I., and Treinen, R., eds., {\em LoCoCo}, volume~29 of {\em
  EPTCS},  11--22.

\bibitem[\protect\citeauthoryear{Bahar \bgroup et al\mbox.\egroup
  }{1993}]{BaharFGHMPS93}
Bahar, R.~I.; Frohm, E.~A.; Gaona, C.~M.; Hachtel, G.~D.; Macii, E.; Pardo, A.;
  and Somenzi, F.
\newblock 1993.
\newblock Algebraic decision diagrams and their applications.
\newblock In {\em {Proceedings of the International Conference Computer-Aided
  Design (ICCAD)}},  188--191.

\bibitem[\protect\citeauthoryear{Biere \bgroup et al\mbox.\egroup
  }{2009}]{DBLP:series/faia/2009-185}
Biere, A.; Heule, M.; van Maaren, H.; and Walsh, T., eds.
\newblock 2009.
\newblock {\em Handbook of Satisfiability}, volume 185 of {\em Frontiers in
  Artificial Intelligence and Applications}. IOS Press.

\bibitem[\protect\citeauthoryear{Boland \bgroup et al\mbox.\egroup
  }{2006}]{Bolandetal06}
Boland, N.; Dom{\'{\i}}nguez{-}Mar{\'{\i}}n, P.; Nickel, S.; and Puerto, J.
\newblock 2006.
\newblock Exact procedures for solving the discrete ordered median problem.
\newblock {\em Computers {\&} {OR}} 33(11):3270--3300.

\bibitem[\protect\citeauthoryear{Boros and Hammer}{2002}]{boros2002pseudo}
Boros, E., and Hammer, P.~L.
\newblock 2002.
\newblock Pseudo-boolean optimization.
\newblock {\em Discrete applied mathematics} 123(1):155--225.

\bibitem[\protect\citeauthoryear{Darwiche and Marquis}{2002}]{compilationMap}
Darwiche, A., and Marquis, P.
\newblock 2002.
\newblock A knowledge compilation map.
\newblock {\em J. Artif. Intell. Res. (JAIR)} 17:229--264.

\bibitem[\protect\citeauthoryear{Darwiche}{1999}]{darwiche1999compiling}
Darwiche, A.
\newblock 1999.
\newblock Compiling knowledge into decomposable negation normal form.
\newblock In {\em IJCAI},  284--289.
\newblock Citeseer.

\bibitem[\protect\citeauthoryear{Darwiche}{2001}]{darwiche2001decomposable}
Darwiche, A.
\newblock 2001.
\newblock Decomposable negation normal form.
\newblock {\em Journal of the ACM (JACM)} 48(4):608--647.

\bibitem[\protect\citeauthoryear{Galand and
  Spanjaard}{2012}]{GalandSpanjaard12}
Galand, L., and Spanjaard, O.
\newblock 2012.
\newblock Exact algorithms for owa-optimization in multiobjective spanning tree
  problems.
\newblock {\em Computers {\&} {OR}} 39(7):1540--1554.

\bibitem[\protect\citeauthoryear{Gebser, Kaminski, and Schaub}{2011}]{aspcud}
Gebser, M.; Kaminski, R.; and Schaub, T.
\newblock 2011.
\newblock aspcud: A linux package configuration tool based on answer set
  programming.
\newblock In Drescher, C.; Lynce, I.; and Treinen, R., eds., {\em LoCoCo},
  volume~65 of {\em EPTCS},  12--25.

\bibitem[\protect\citeauthoryear{Janota \bgroup et al\mbox.\egroup
  }{2012}]{packup}
Janota, M.; Lynce, I.; Manquinho, V.~M.; and Marques-Silva, J.
\newblock 2012.
\newblock Packup: Tools for package upgradability solving.
\newblock {\em JSAT} 8(1/2):89--94.

\bibitem[\protect\citeauthoryear{Karp}{1972}]{karp1972reducibility}
Karp, R.~M.
\newblock 1972.
\newblock {\em Reducibility among combinatorial problems}.
\newblock Springer.

\bibitem[\protect\citeauthoryear{Kimmig, {Van den Broeck}, and {De
  Raedt}}{2012}]{Kimmigetal12}
Kimmig, A.; {Van den Broeck}, G.; and {De Raedt}, L.
\newblock 2012.
\newblock Algebraic model counting.
\newblock Technical Report 1211.4475, ArXiv, Cornell University Library.

\bibitem[\protect\citeauthoryear{Li and Many{\`a}}{2009}]{MAXSAT}
Li, C.~M., and Many{\`a}, F.
\newblock 2009.
\newblock Maxsat, hard and soft constraints.
\newblock In Biere et~al. \shortcite{DBLP:series/faia/2009-185}.
\newblock  613--631.

\bibitem[\protect\citeauthoryear{Mancinelli \bgroup et al\mbox.\egroup
  }{2006}]{mancinelli2006managing}
Mancinelli, F.; Boender, J.; Di~Cosmo, R.; Vouillon, J.; Durak, B.; Leroy, X.;
  and Treinen, R.
\newblock 2006.
\newblock Managing the complexity of large free and open source package-based
  software distributions.
\newblock In {\em Automated Software Engineering, 2006. ASE'06. 21st IEEE/ACM
  International Conference on},  199--208.
\newblock IEEE.

\bibitem[\protect\citeauthoryear{Moulin}{1991}]{moulin1991axioms}
Moulin, H.
\newblock 1991.
\newblock {\em Axioms of cooperative decision making}.
\newblock Cambridge University Press.

\bibitem[\protect\citeauthoryear{Nemhauser and Wolsey}{1988}]{MILP}
Nemhauser, G.~L., and Wolsey, L.~A.
\newblock 1988.
\newblock {\em Integer and combinatorial optimization}.
\newblock Wiley interscience series in discrete mathematics and optimization.
  Wiley.

\bibitem[\protect\citeauthoryear{Ogryczak and
  {\'S}liwi{\'n}ski}{2003}]{ogryczak2003solving}
Ogryczak, W., and {\'S}liwi{\'n}ski, T.
\newblock 2003.
\newblock On solving linear programs with the ordered weighted averaging
  objective.
\newblock {\em European Journal of Operational Research} 148(1):80--91.

\bibitem[\protect\citeauthoryear{Pipatsrisawat and
  Darwiche}{2008}]{PipatsrisawatDarwiche08}
Pipatsrisawat, K., and Darwiche, A.
\newblock 2008.
\newblock New compilation languages based on structured decomposability.
\newblock In {\em Proc. of AAAI'08},  517--522.

\bibitem[\protect\citeauthoryear{Rosenberg}{1975}]{rosenberg1975reduction}
Rosenberg, I.
\newblock 1975.
\newblock Reduction of bivalent maximization to the quadratic case.
\newblock {\em Cahiers du Centre d'{\'e}tudes de Recherche Operationnelle}
  17:71--74.

\bibitem[\protect\citeauthoryear{Roussel and Manquinho}{2009}]{PBO}
Roussel, O., and Manquinho, V.~M.
\newblock 2009.
\newblock Pseudo-boolean and cardinality constraints.
\newblock In Biere et~al. \shortcite{DBLP:series/faia/2009-185}.
\newblock  695--733.

\bibitem[\protect\citeauthoryear{Sanner and McAllester}{2005}]{SannerM05}
Sanner, S., and McAllester, D.~A.
\newblock 2005.
\newblock Affine algebraic decision diagrams ({AADD}s) and their application to
  structured probabilistic inference.
\newblock In {\em {Proceedings of the Nineteenth International Joint Conference
  on Artificial Intelligence (IJCAI)}},  1384--1390.

\bibitem[\protect\citeauthoryear{Syrjänen}{1999}]{debian-asp-99}
Syrjänen, T.
\newblock 1999.
\newblock A rule-based formal model for software configuration.
\newblock Master's thesis, Helsinki University of Technology.

\bibitem[\protect\citeauthoryear{Tucker \bgroup et al\mbox.\egroup
  }{2007}]{opiumPaper}
Tucker, C.; Shuffelton, D.; Jhala, R.; and Lerner, S.
\newblock 2007.
\newblock Opium: Optimal package install/uninstall manager.
\newblock In {\em ICSE},  178--188.
\newblock IEEE Computer Society.

\bibitem[\protect\citeauthoryear{Wilson}{2005}]{Wilson05}
Wilson, N.
\newblock 2005.
\newblock Decision diagrams for the computation of semiring valuations.
\newblock In {\em {Proceedings of the Nineteenth International Joint Conference
  on Artificial Intelligence (IJCAI)}},  331--336.

\bibitem[\protect\citeauthoryear{Yager}{1988}]{Yager88}
Yager, R.~R.
\newblock 1988.
\newblock On ordered weighted averaging aggregation operators in multicriteria
  decisionmaking.
\newblock {\em {IEEE} Transactions on Systems, Man, and Cybernetics}
  18(1):183--190.

\end{thebibliography}

\vfill
\pagebreak
\section{Appendix: Proof Sketches}

\begin{proof}[Proposition \ref{prop:obvious}]
Associate in polynomial time with any formula $\phi$ from $L$ the hard constraint $\phi$ and the empty
weighted base. $\phi$ is consistent if and only if the output of the optimization
algorithm is not "no solution".
\end{proof}

\begin{proof}[Proposition \ref{prop:pi:np}]
  Every positive Krom formula (i.e., a \cnf\ formula $\phi$
  consisting of binary clauses, where every variable occurs in $\phi$
  only positively) can be turned into an equivalent \pi\
  formula in polynomial time where all literals are positive.
   The {\sf NP}-hardness of optimization in this case comes from the
  {\sf NP}-hardness of the minimal hitting set problem, where the
  cardinality of the sets is 2. Let $C$ be a set of sets $c_i$
  (where $|c_i| = 2$) whose elements are included in a reference set
  $E$. A hitting set of $C$ is a set $h$ of elements of $E$ such that
  $E \cap c_i \neq \emptyset$. Determining whether a hitting set $h$
  of $C$ containing at most $k$ elements exists or not has been proved
  {\sf NP}-complete ~\cite{karp1972reducibility}. Clearly, such a
  hitting set $h$ exists if and only if any optimal solution $\omega^*$ of the optimization
  problem given by the hard constraint $\phi = \bigwedge_{c_i \in C} \vee_{x \in c_i}$
  (a positive Krom formula) and the weighted base $\{(x, 1) \mid x \in \PS\}$
  (which is in $\mathcal{L}_+^+$) is such that $f(\omega^*) \leq q$
  when $\oplus = \Sigma$, and $f(\omega^*) $ contains at most $k$ ones
  when $\oplus = \mathit{leximax}$.
\end{proof}

\begin{proof}[Lemma \ref{lemma:optFromModGen}]
  Given a model generator $\omega_V$ of $\phi$ and a linear weighted base
  $\{(l_i, w_i) \mid i \in \{1, \ldots, n\}\}$, it is easy to compute an optimal model of $\phi$ which extends $\omega_V$: 
  consider successively every literal $l_i$ the variable $v_i$ of which is not assigned by $\omega_V$;
  if $w_i < 0$, then add $(v_i, 0)$ to $\omega_V$ if $l_i$ is a negative literal and
  $(v_i, 1)$ to $\omega_V$ otherwise; if $w_i \geq 0$, then add $(v_i, 0)$ to $\omega_V$ if $l_i$ is a positive literal and
  $(v_i, 1)$ to $\omega_V$ otherwise.
\end{proof}

\begin{proof}[Lemma \ref{lemma:leafModGen}]
  There are four kinds of \nnf\ formulae of size $1$: if $\phi = \top$,
  then the models of $\phi$ are the extensions of the model generator
  $\{\}$; if $\phi = \bot$, then $\phi$ has no model, so no model
  generator of $\phi$ exists.  If $\phi = x \in \PS$, then the models
  of $\phi$ are the extensions of the model generator $\{(x,1)\}$;
  finally, if $\phi = \neg x$ with $x \in \PS$, then the models of
  $\phi$ are the extensions of the model generator $\{(x,0)\}$.
\end{proof}

\begin{proof}[Lemma \ref{lemma:andModGen}]
  If one of the $\phi_i$ is inconsistent, then it has no model
  generator, and as a consequence, $\phi$ has no model generator. In
  the remaining case, thanks to the decomposability property, the
  union of the model generators of all the children $\phi_i$ is a model
  generator of $\phi$.
\end{proof}

\begin{proof}[Lemma \ref{lemma:orModGen}]
The inconsistency of $\phi$ can be easily detected since 
$\phi$ is inconsistent if and only if all the children $\phi_i$ of $\phi$
are inconsistent, hence do not have model generators. 
In the remaining case, since $\phi$ is a disjunction, the
models of $\phi$ are the union of the models of its children. 
Hence, every optimal model of $\phi$ given $f$ is an optimal model of at
least one of its children. Altogether, this
implies that at least one of the children $\phi_i$ of $\phi$ is associated
with a model generator which can be extended to an
optimal model of $\phi$ given $f$. 
Thanks to Lemma \ref{lemma:optFromModGen}, one computes in polynomial time
an optimal model $\omega_i^*$ of each $\phi_i$ from the associated generator; it is then enough
to compare the values $f(\omega_i^*)$ to determine an $\omega_i^*$ which is an optimal model
of $\phi$, and to select the corresponding model generator.
\end{proof}

\begin{proof}[Proposition \ref{prop:dnnf:p}]
 Direct by structural induction on $\phi$ given Lemmas
  \ref{lemma:optFromModGen},\ref{lemma:leafModGen},\ref{lemma:andModGen}
  and \ref{lemma:orModGen}.
\end{proof}

\begin{proof}[Proposition \ref{prop:owa}]
  We point out a polynomial-time reduction from \opt[$L$,$\mathcal{Q_+}$,$\Sigma$] to
  \opt[$L$,$\mathcal{L}_+$,$\mathit{OWA}_W$] for some $W$.  The existence
  of such a reduction is enough to get the result given Proposition \ref{prop:nnf:q+0:np}.
  Let us recall that given a $s$-vector of real numbers $W = (p_1, \ldots, p_s)$ such that $\Sigma_{i=1}^n p_i = 1$, 
  $\mathit{OWA}_W$ is the mapping associating with any $s$-vector of real numbers 
  $V = (v_1, \ldots, v_s)$ the real number given by   $\mathit{OWA}_W(V) = \Sigma_{i=1}^s p_i  v_{\sigma(i)}$
  where $\sigma$ is the permutation of $\{1, \ldots, n\}$ such that $v_{\sigma(1)} \geq \ldots \geq v_{\sigma(n)}$.
  Consider now any $\mathcal{Q_+}$ representation of an objective function $f$ of the form of the weighted base 
  $\{(l_{i,1} \wedge l_{i,2}, w_i) \mid i \in \{1, \ldots, n\}\}$. Let $K = \mathit{max}(\{w_i, i=\{1,\ldots,n\}\})+1$.
  We associate in linear time with this weighted base the following weighted base
  $\{(l_{i,1}, n[K(n-i+1)+w_i]), (l_{i,2}, n[K(n-i+1)+w_i]), (\sim l_{i,1}, nK(n-i+1)),  (\sim l_{i,2}, nK(n-i+1)) \mid i \in \{1, \ldots, n\}\}$,
  and we consider the $\mathit{OWA}_W$ aggregator given by the $4n$-vector of real numbers
  $W = (0, \frac{1}{n}, 0, \frac{1}{n}, \ldots, 0, \frac{1}{n}, 0, 0, \ldots, 0)$ starting with the sequence $0, \frac{1}{n}$ repeated $n$ times and followed
  by a sequence of $2n$ zeroes.
  Let $g$ be the objective function represented by such a base when this aggregator is considered.
  We want to prove that for any pair of interpretations $\omega$, $\omega'$, we have 
  $f(\omega) \leq f(\omega')$ if and only if   $g(\omega) \leq g(\omega')$.
  Consider any interpretation $\omega$ and for any $i \in \{1, \ldots, n\}$ the four weighted formulae
  $(l_{i,1}, n[K(n-i+1)+w_i]), (l_{i,2}, n[K(n-i+1)+w_i]), (\sim l_{i,1}, nK(n-i+1)),  (\sim l_{i,2}, nK(n-i+1))$.
  By construction for each $j \in \{1, 2\}$, $\omega$ satisfies $l_{i,j}$ precisely when it does not
  satisfy $\sim l_{i,j}$. Observe also that $nK(n-i+1)$ (and a fortiori $n[K(n-i+1)+w_i]$) is strictly greater than 
  $n[K(n-(i+1)+1)+w_{i+1}]$ for each $i \in \{1, \ldots, n-1\}$ since $K$ is strictly greater than each $w_{i+1}$.
  Accordingly, the vector of values sorted in non-increasing way induced by $\omega$ and the
  base representing $g$ has the form $(a_1, b_1, \ldots, a_n, b, n, 0, \ldots, 0)$ where for
  each $i \in \{1, \ldots, n\}$, $a_i = b_i = n[K(n-i+1)+w_i]$ when $\omega$ satisfies $l_{i,1} \wedge l_{i,2}$,
  $a_i = n[K(n-i+1)+w_i]$ and $b_i = nK(n-i+1)$ when $\omega$ satisfies $\sim l_{i,1} \wedge l_{i,2}$ or
  $\omega$ satisfies $l_{i,1} \wedge \sim l_{i,2}$, and $a_i = b_i = nK(n-i+1)$ when $\omega$ satisfies 
  $\sim l_{i,1} \wedge \sim l_{i,2}$. Thus, for each $i \in \{1, \ldots, n\}$, $b_i =  n[K(n-i+1)+w_i]$
  when $\omega$ satisfies $l_{i,1} \wedge l_{i,2}$ (otherwise $b_i = nK(n-i+1)$).  Since the vector $W$
  starts with the sequence $0, \frac{1}{n}$ repeated $n$ times, the value of $g(\omega)$ is independent from the $a_i$ 
  and depends only of the $b_i$. More precisely, we have $g(\omega) = (\Sigma_{i=1}^n w_i.( l_{i,1} \wedge l_{i,2})(\omega))
  + (\Sigma_{i=1}^n K(n-i+1)) = f(\omega) + K\frac{n(n+1)}{2}$. Since $K\frac{n(n+1)}{2}$ is a constant term independent
  from $\omega$, we get the expected result.
%
\end{proof}

\begin{proof}[Proposition \ref{prop:nnf:q+0:np}]~\\
\vspace{-4mm}
\begin{itemize}
\item \opt[$L$, $\mathcal{Q}_+$, $\oplus$]. Consider the following decision problem. 
Given a finite set $S = \{\gamma1, \ldots, \gamma_n\}$ of terms of size 2 and an integer $k$,
one wants to determine whether there exists an interpretation satisfying at least $k$ terms of $S$. This problem is known as {\sf NP}-complete. 
Now, we can associate in polynomial time with any $S$ and $k$ the instance of \opt[$L$, $\mathcal{Q}_+$, $\oplus$] 
given by $\phi = \top$, and the weighted base $\{(\sim l_{i,1} \wedge l_{i,2}, 1), 
(l_{i,1} \wedge \sim l_{i,2}, 1), (\sim l_{i,1} \wedge \sim l_{i,2}, 1) \mid \gamma_i = l_{i,1} \wedge l_{i,2} \in S\}$. By construction, the optimal solution $\omega^*$ 
of the instance of \opt[$L$, $\mathcal{Q}_+$, $\oplus$] is such that $f(\omega^*) \leq n-k$ when $\oplus = \Sigma$ and $f(\omega^*)$ contains at least $k$ zeroes
(hence at most $n-k$ ones) when $\oplus = \mathit{leximax}$ if and only if there exists an interpretation (namely, $\omega^*$) satisfying at least $k$ terms of $S$.
This proves that \opt[$L$, $\mathcal{Q}_+$, $\oplus$] is {\sf NP}-hard for $\oplus = \Sigma$ and $\oplus = \mathit{leximax}$.

\item \opt[$L$, $\mathcal{Q}^+$, $\oplus$]. Consider the the minimal hitting set problem, where the cardinality of the sets is 2,
as in the proof of Proposition \ref{prop:pi:np}. With each element $x$ of the reference set $E = \bigcup_{c_i \in C} c_i$ of cardinality $n$ 
corresponds a propositional variable $x$ meaning that $x$ is not selected in the hitting set. We can associate in polynomial time with $C$, the given collection 
of $m$ subsets $c_i$ of $E$ and the constant $k$, the instance of \opt[$L$, $\mathcal{Q}^+$, $\oplus$] given by $\phi = \top$ and
the weighted base $\{(x_{i,1} \wedge x_{i,2}, 2), (x_{i,1}, -1), (x_{i,2}, -1) \mid c_i = \{x_{i,1}, x_{i,2}\} \in C\}$ in which
every occurrence of a multi-occurrent pair is removed but one (so that the resulting multi-set actually is a set containing $m$+$n$ weighted formulae).
Observe that for each $c_i = \{x_{i,1}, x_{i,2}\} \in C$ and every interpretation $\omega$, the "contribution" of the 
restriction over $\{x_{i,1}, x_{i,2}\}$ of $\omega$ to $f(\omega)$ conveyed by the weighted formulae $(x_{i,1} \wedge x_{i,2}, 2), (x_{i,1}, -1), (x_{i,2}, -1) $
associated with $c_i$ consists of the aggregation of values $2$, $-1$ and $0$: when both variables $x_{i,1}$ and $x_{i,2}$ are set to $1$ (i.e., no element of $c_i$
is selected so that the resulting set will not be a hitting set) the obtained values are $2$, $-1$, $-1$, when one of them only is set to $1$, 
the obtained values are $0$, $0$, $-1$, and finally when both variables are set to $0$, the obtained values are $0$, $0$, $0$.
By construction the interpretation assigning every variable to $0$ always is a hitting set of $C$ (this hitting set is equal to $E$).
When an interpretation corresponds to a hitting set of $C$, the values to be aggregated are only among $0$ and $-1$.
The optimal solution $\omega^*$ of the instance of \opt[$L$, $\mathcal{Q}_+$, $\mathit{leximax}$] is such that $f(\omega^*)$ contains 
as many $-1$ as possible (and no $2$). Thus, $C$ has a hitting set of size at most $k$ if and only if $n + f(\omega^*) \leq k$
(when $\oplus = \Sigma$) and $f(\omega^*)$ contains at least $n-k$ $-1$ (when $\oplus = \mathit{leximax}$).
\end{itemize}
\end{proof}

\begin{proof}[Proposition \ref{prop:dnf:p}]
Let us consider a general representation of the objective function, with positive literals and 
positive weights $\{(\phi_i, w_i) \mid i \in \{1, \ldots, n\}\}$. When each $w_i$ is positive,
since a minimal solution is targeted and given the aggregators which are considered, an interpretation which does not satisfy $\phi_i$ (hence leading to
a weight $0$) is always preferred to an interpretation satisfying $\phi_i$ and leading to a weight $w_i \geq 0$.
Furthermore, positive formulae $\phi_i$ are monotone: if an interpretation $\omega$ is a model of $\phi_i$
then every interpretation which coincides with $\omega$ on the variables set to $1$ by $\omega$ also is
a model of $\phi_i$.

Let $\phi = \bigvee_{j=1}^m t_j$ be a \dnf\ formula. 
A polynomial-time algorithm for \opt[\dnf, $\mathcal{G}^+_+$, $\oplus$] is as follows. 
If $\phi$ contains no consistent term, then return "no solution".
Otherwise, for each consistent term $t_j$ of $\phi$, let us define $\omega_{t_j}$ as the interpretation which 
satisfies $t_j$ and sets to $0$ every variable which does not occur in $t_j$. By construction, among the
interpretations $\omega$ satisfying $t_j$, if  $\omega$ does not satisfy $\phi_i$, then necessarily this is also the case of
$\omega_{t_j}$. Thus $\omega_{t_j}$ is an optimal solution among the models of $t_j$.
Since the models of $\phi$ are exactly the interpretations satisfying at least one $t_j$, it is enough to 
compare in a pairwise fashion the values $f(\omega_{t_j})$ for each $t_j$ of $\phi$
to determine an $\omega_{t_j}$ which is optimal for $\phi$, and finally to return it.
\end{proof}

\begin{proof}[Proposition \ref{prop:obddpi:np}]
  We have proved (Proposition \ref{prop:nnf:q+0:np})
  that \opt[$L$, $\mathcal{Q}_+$, $\oplus$] is {\sf NP}-hard for
  $\oplus = \Sigma$ and for $\oplus = \mathit{leximax}$, even when the hard constraint 
  $\phi = \top$ belongs to $L$, which is the case when $L$ =
  \oobdd\ or $L$ = \pi.  In such a case, we point out a polynomial-time reduction from
  \opt[\oobdd, $\mathcal{Q}_+$, $\oplus$] to \opt[\oobdd, $\mathcal{Q}^+_+$, $\oplus$]  
  and from \opt[\pi, $\mathcal{Q}_+$, $\oplus$] to
  \opt[\pi, $\mathcal{Q}^+_+$] problem. It consists in associating in
  linear time with any $\mathcal{Q}_+$ representation a
  $\mathcal{Q}^+_+$ representation obtained by replacing every
  negative literal $\neg x$ occurring in it by a new variable $n_x$
  ($X$ denotes the set of variables occurring in such literals). Then
  one replaces $\phi = \top$ by an \oobdd\ (resp. a \pi)
  representation of $\bigwedge_{x \in X} (\neg x \Leftrightarrow
  n_x)$. The point is that an \oobdd\ (resp. a \pi) representation of
  $\bigwedge_{x \in X} (\neg x \Leftrightarrow n_x)$ can be computed
  in time linear in the input size.  Finally, by construction, both
  optimization problems have the same optimal solutions (once
  projected on the initial variables).
\end{proof}

\begin{proof}[Proposition \ref{prop:cdco:p:p}]
Let $\phi \in L$ and $\{(t_i, w_i)\mid i=\{1,\ldots,n\}\}$ the
$\mathcal{P}$ representation of the objective function. There are
$2^n$ sets of the form $\{t_i^* \mid i=\{1,\ldots,n\}\}$ where
each $t_i^*$ is either $t_i$ or the clause equivalent to $\neg t_i$
and obtained as the disjunction of the negations of the literals occurring in $t_i$.
With each of those sets corresponds a $n$-vector of reals $(v_1, \ldots, v_n)$
where each $v_i$ ($i \in \{1, \ldots, n\}$) is equal to $w_i$ when $t_i$ is in the set,
and to $0$ otherwise.
When $n$ is bounded, the number of those sets is bounded as well.
Furthermore, when interpreted conjunctively, each set can be viewed as a \cnf\ formula
$\alpha$ which contains at most $n$ clauses of size $> 1$. 
Each of the clauses in $\alpha$ contains at most $m$ literals (where $m$  is
the cardinality of $\PS$). As a consequence, each $\alpha$ can be turned in
time $O(m^n)$ into an equivalent \dnf\ formula $\alpha'$, which contains 
at most $O(m^n)$ consistent terms. When $L$ satisfies \cd\ and \co, we can easily 
check whether $\alpha' \wedge \phi$ is consistent by determining whether there
exists a consistent term $t$ in $\alpha'$ so that $\phi \mid t$ is consistent.
In the enumeration process, every $\alpha'$ such that 
$\alpha' \wedge \phi$ is inconsistent is simply skipped. For each remaining $\alpha'$
one can easily compute the value $f(\omega_{\alpha'})$ where $\omega$ is any model of $\alpha'$,
hence any model of $\alpha$. Indeed, by construction, it is equal to the aggregation
function $\oplus$ applied to the $n$-vector of reals $(v_1, \ldots, v_n)$ associated with
$\alpha$. Hence at each step of the enumeration it is enough to memorize one of the
$\alpha'$ encountered so far leading to a minimal value $f(\omega_{\alpha'})$.
Once all the $\alpha$ have been considered, if no $\alpha$ such that
$\alpha' \wedge \phi$ is consistent has been found, the algorithm returns "no solution";
in the remaining case, an "optimal" $\alpha'$ has been stored and 
every interpretation extending a consistent term $t$ in $\alpha'$ so that $\phi \mid t$ is consistent
is an optimal solution.
\end{proof}


\begin{proof}[Proposition \ref{prop:obdd:obdd:fpt}]
One takes advantage of a linearization process here.
Basically the approach consists in replacing each $\phi_i$ which is not a literal by a new variable $n_{\phi_i}$
and in replacing the hard constraint $\phi$ by an \oobdd\ representation equivalent to 
$\phi \wedge \bigwedge_{i=1}^n (n_{\phi_i} \Leftrightarrow \phi_i)$. The key point is that this
\oobdd\ representation can be computed in $\mathcal{O}|\phi|. (2\mathit{max}_{i=1}^n |\phi_i|+1)^{n-1})$ time
because the conjunction of a bounded number of \oobdd\ representations can be computed in polynomial time as an \oobdd\ 
representation and an \oobdd\ representation of $n_{\phi_i} \Leftrightarrow \phi_i$ can be
computed in time polynomial in the size of the \oobdd\ formula $\phi_i$, whatever $<$. The proof for \tDnnf\ is similar (the point is that \tDnnf\ satisfies the same bounded conjunction transformation as \oobdd).
\end{proof}

\begin{proof}[Proposition \ref{prop:l:g+0:nfpt}]
We consider the problem of determining the consistency of a \cnf\ formula of the form $\psi^+ \wedge \psi^-$
where every clause of $\psi^+$ is positive (i.e., it consists of positive literals only) while every clause of $\psi^-$ is 
negative (i.e., it consists of negative literals only). It is well-known that this restriction of {\sc cnf-sat} is {\sf NP}-complete.
\begin{itemize}
\item \opt[$L$, $\mathcal{G}^+$, $\oplus$]. We can associate in polynomial time with every \cnf\ formula of the form $\psi^+ \wedge \psi^-$ the instance
of \opt[$L$, $\mathcal{G}^+$, $\oplus$] given by $\phi = \top$ and the weighted base $\{(\psi^+, -1)(\neg \psi^-, 1)\}$ 
(note that both $\psi^+$ and $\neg \psi^-$ belongs to $\mathcal{G}^+$).  The point is that $\psi^+ \wedge \psi^-$ is consistent
if and only if the optimal solution $\omega^*$ of \opt[$L$, $\mathcal{G}^+$, $\oplus$] which is computed 
satisfies $f(\omega^*) = -1$ when $\oplus = \Sigma$ and $f(\omega^*) = (0, -1)$ when $\oplus = \mathit{leximax}$.
\item \opt[$L$, $\mathcal{G}_+$, $\oplus$]. We can associate in polynomial time with every \cnf\ formula of the form $\psi^+ \wedge \psi^-$ the instance
of \opt[$L$, $\mathcal{G}_+$, $\oplus$] given by $\phi = \top$ and the weighted base $\{(\neg \psi^+, 1)(\neg \psi^-, 1)\}$ 
The point is that $\psi^+ \wedge \psi^-$ is consistent if and only if the optimal solution $\omega^*$ of \opt[$L$, $\mathcal{G}_+$, $\oplus$] which is computed 
satisfies $f(\omega^*) = 0$ when $\oplus = \Sigma$ and $f(\omega^*) = (0, 0)$ when $\oplus = \mathit{leximax}$.
\end{itemize}
\end{proof}


\begin{proof}[Proposition \ref{corol:obddpi:g++:nfpt}]
The result is a consequence of the fact that \opt[$L$, $\mathcal{G}_+$, $\oplus$] is {\sf NP}-hard
for $\oplus = \Sigma$ and $\oplus = \mathit{leximax}$ even when the hard constraint $\phi$ is $\top$
and $n \geq 2$ (cf. Proposition \ref{prop:l:g+0:nfpt}).
Similarly to the reduction used in the proof of Proposition  \ref{prop:obddpi:np}, 
one associates in linear time with any $\mathcal{G}_+$ representation a
$\mathcal{G}^+_+$ representation obtained by replacing every
 negative literal $\neg x$ occurring in it by a new variable $n_x$
 ($X$ denotes the set of variables occurring in such literals). Then
 one considers as the new hard constraint an \oobdd\ (resp. a \pi)
  representation $\phi$ of $\bigwedge_{x \in X} (\neg x \Leftrightarrow
  n_x)$. The point is that an \oobdd\ (resp. a \pi) representation of
  $\bigwedge_{x \in X} (\neg x \Leftrightarrow n_x)$ can be computed
  in time linear in the input size. By construction, both
  optimization problems have the same optimal solutions (once
  projected on the initial variables).
%
\end{proof}

\end{document}